\documentclass{article}

\PassOptionsToPackage{numbers, compress}{natbib}

\usepackage[preprint]{neurips_2024}

\usepackage[utf8]{inputenc} 
\usepackage[T1]{fontenc}    
\usepackage{hyperref}       
\usepackage{url}            
\usepackage{booktabs}       
\usepackage{amsfonts}       
\usepackage{nicefrac}       
\usepackage{microtype}      
\usepackage{xcolor}         
\usepackage{graphicx}

\title{Scaling Laws for Neural Material Models}

\author{%
  Akshay Trikha$^a$ \And Kyle Chu$^a*$ \And Advait Gosai$^a*$ \AND Parker Szachta$^a*$ \And Eric Weiner$^a$ \AND \textit{$^a$University of California, Berkeley} \AND \textit{$*$Equal contribution}\\
}

\begin{document}

\maketitle


\begin{abstract}
    Predicting material properties is crucial for designing better batteries, semiconductors, and medical devices. Deep learning helps scientists quickly find promising materials by predicting their energy, forces, and stresses. Companies scale capacities of deep learning models in multiple domains, such as language modeling, and invest many millions of dollars into such models. Our team analyzes how scaling training data (giving models more information to learn from), model sizes (giving models more capacity to learn patterns), and compute (giving models more computational resources) for neural networks affects their performance for material property prediction. In particular, we trained both transformer and EquiformerV2 neural networks to predict material properties. We find empirical scaling laws for these models: we can predict how increasing each of the three hyperparameters (training data, model size, and compute) affects predictive performance. In particular, the loss $L$ can be measured with a power law relationship $L = \alpha \cdot N^{-\beta}$, where $\alpha$ and $\beta$ are constants while $N$ is the relevant hyperparameter. We also incorporate command-line arguments for changing training settings such as the amount of epochs, maximum learning rate, and whether mixed precision is enabled. Future work could entail further investigating scaling laws for other neural network models in this domain, such as GemNet and fully connected networks, to assess how they compare to the models we trained.
\end{abstract}

\section{Section 1}
\subsection{Introduction}



Predicting material properties with high accuracy is a fundamental challenge in materials science with significant implications for drug discovery and materials discovery for better nuclear materials, batteries, semiconductors, and medical devices. Traditionally, researchers have relied on density functional theory (DFT) to perform these computations. Though DFT is largely reliable it is computationally expensive as it involves solving approximations of Schrodinger’s equation and scales approximately on the order of $O(n^3)$, where $n$ is the number of electrons in the system. Many have tried to bypass the time complexity bottleneck of DFT by training deep learning models to directly predict the results of DFT simulations given input structures. This shift has been enabled by the exponential growth of DFT datasets in recent years, which now support the training of large models with promising accuracy. 

Composed of 118 million structure-property pairs, the Open Materials 2024 (OMat24) dataset is one of the largest public collections for training deep learning models in materials science \cite{OMat24}. It was sampled from the Alexandria PBE dataset \cite{Alexandria} and contains a diverse set of inorganic crystal structures labeled with energy, force, and stress information from DFT calculations. OMat24 emphasizes non-equilibrium configurations and compositional diversity, enabling improved generalization in models trained on the dataset. Sampling strategies include Boltzmann-weighted rattled structures, ab initio molecular dynamics (AIMD), and relaxed rattled structures. At the time of the OMat24 publication, the state-of-the-art model on the Matbench Discovery benchmark was an EquiformerV2 model pre-trained on OMat24 and finetuned on the MPTrj and sAlex datasets.

To better understand the limits of neural network-based models in this domain, we investigate the scaling behavior of two architectures: an encoder-only transformer \cite{ADiT} and EquiformerV2 \cite{EquiformerV2}. The former being an unconstrained model and the latter being a very physically constrained model, allowing us to explore whether explicit enforcement of physical symmetries improves performance or if unconstrained models learn these properties implicitly at a large enough scale.

Our study uses the OMat24 dataset to investigate how varying training data size, model size (measured in parameters), and compute (measured in FLOPs) impact predictive performance when one variable is fixed and the others are varied. By empirically testing scaling laws across model sizes ranging from $10^2$ to nearly $10^9$ parameters, we aim to provide a heuristic for selecting architectures and training strategies as computational resources and data availability increase. These results will help identify when scaling yields diminishing returns for our domain of material property prediction, and optimize resource allocation.

\subsection{Related Work}

A major topic of discussion in the materials science industry involves assessing the properties of materials. Our project aims to assess how worthwhile it is to scale neural models on such data (see paragraph $2.2.1$ below).

\subsubsection{Scaling Laws}

Deep learning models often exhibit predictable improvement when they are trained on increasing amounts of data with increasing learnable parameters. Early experiments from Hestness et al. in 2017 first hinted at scaling behavior by showing that in the machine translation, language modeling, image classification, and speech recognition domains the generalization error drops as a power law as training set size and model capacity increase \cite{BaiduScaling}. Three years later while training general purpose language models, Kaplan et al. reported a similar trend that their models' 
cross-entropy loss also decreased smoothly with increasing model parameters and data \cite{ScalingLawsForLLMs}. These studies established that deep learning model performance can be reliability predicted by modeling their validation loss as a power law. However, they focus on the domain of language modeling, while our project focuses on materials data. Using some of these models, our project aims to assess the extent to which having physical assumptions baked into models affects predictive performance.


\subsubsection{Graph-Based Models and Equivariant Architectures for Materials Science}

In parallel to progress in language modeling, there have also been advancements in modeling atomic interactions using deep learning. In 2017, Gilmer et al. showed that using graph structures to represent atomic systems turned out to be a useful inductive bias \cite{NeuralMessagePassing}. Since then graph neural networks (GNNs) have become the default backbone for models that learn to predict atomic behavior. Building on the message passing framework, Schütt et al. created SchNet, a model which took interatomic distances as input features which ensured that its learned representations are invariant to translation and rotation \cite{SchNet}. By jointly predicting total energy and atomic forces with a featurization that respects fundamental symmetries, SchNet achieved state-of-the-art accuracy on the QM9 dataset \cite{QM9}. Inspired by the success of physics informed models, Batzner et al. trained NequIP which was the first to use features that were both rotationally and translationally E(3) equivariant. Their research concluded that incorporating equivariant features significantly reduces data requirements for modeling potential energy surfaces \cite{NequIP}. EquiformerV2, the successor to Equiformer, from Schmidt et al. went one step further than NequIP in learning E(n) equivariant features by modifying the standard attention mechanism \cite{EquiformerV2}. The current state-of-the-art model for interatomic potentials is EScAIP which does not strictly enforce rotational symmetry in its learned features and achieves its performance by using a novel neighbor-level multi-head self-attention mechanism within its GNN. Qu et al. emphasize the importance of scalability in the long run by demonstrating that their slightly modified attention mechanism improves both the scalability and expressivity of their model.

\subsection{Conclusion}
Our project aims to explore how neural models scale as we change both the size of training data and the number of model parameters on material datasets. We plan to experiment with various architectures, such as fully connected networks, transformers, and graph neural networks, to see which model type can best predict molecular energy, forces, and stresses on the OMat24 dataset. Another key question to investigate is whether larger models can learn symmetries on their own, or explicit equivariance constraints are necessary to increase performance.

Understanding these scaling trends is critical for researchers to identify when increasing model size or adding more data yields meaningful improvements and when diminishing returns set in. This will guide the design of future models balancing performance and computational efficiency, particularly in fields like drug discovery, catalyst design, and materials engineering, where accurate molecular predictions can significantly accelerate scientific progress. By establishing a power-law relationship, we aim to provide a heuristic for large scale training runs where the compute and cost is substantial, making it only feasible to train once with a high level of accuracy and predictability.

\section{Section 2}

\subsection{Experiment Structure}
Our project focuses on predicting data from the Open Materials 2024 dataset (OMat24), a recently released dataset in the materials science domain \cite{OMat24}. The inputs to our models are the atomic numbers and atomic positions of atoms in a material; the outputs are predicted material energy, atomic forces, and material stresses. To our knowledge, we are assessing scaling on a unique set of models.

\subsubsection{Data Analysis}
To ensure that model designs are able to significantly learn from training data, we assess models on small amounts of data to analyze whether the models are able to overfit on data. This occurs when training loss (a measure of how incorrect a model performs on data that the model has seen) is very low but validation loss (a measure of how incorrect a model performs on data that the model has not seen) is relatively high. Overfitting occurs when a model fits too tightly to training data and cannot generalize. If overfitting can occur for a model, that model is able to learn from data. Once we are able to overfit our models, we run two types of experiments: (1) provide a constant amount of training data while varying the model sizes, and (2) provide a constant model size while varying the amount of training data. From the resulting loss curves, we fit power law relationships to these curves, representing empirical scaling laws \cite{ScalingLawsForLLMs}.

\subsection{Methods}

\subsubsection{Summary Statistics}
\begin{figure}
\centering
\includegraphics[scale=0.12]{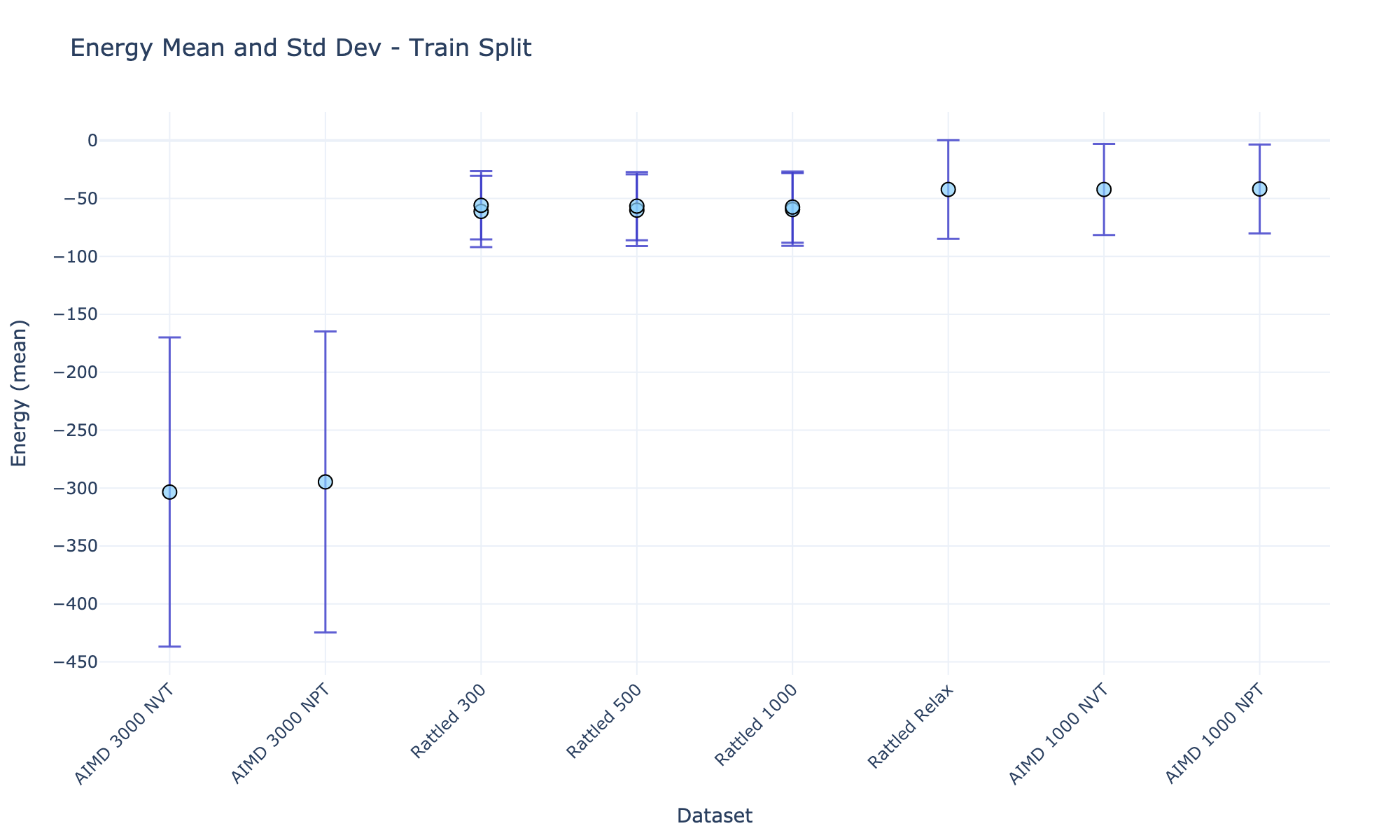}\includegraphics[scale=0.12]{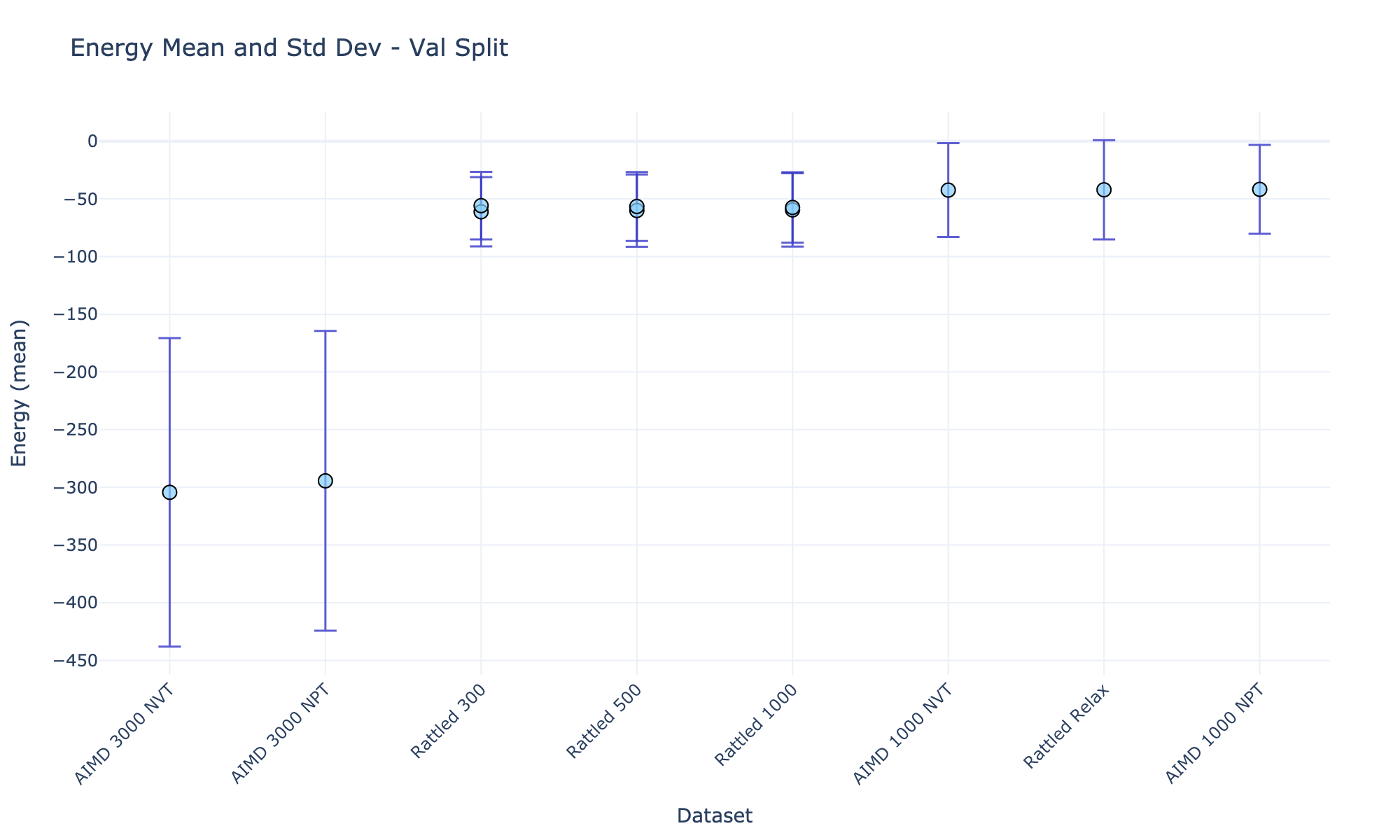}
\caption{We computed mean and standard deviations of energy for datasets and splits.}
\label{fig:data_analysis_energy}
\end{figure}

\begin{figure}
\centering
\includegraphics[scale=0.12]{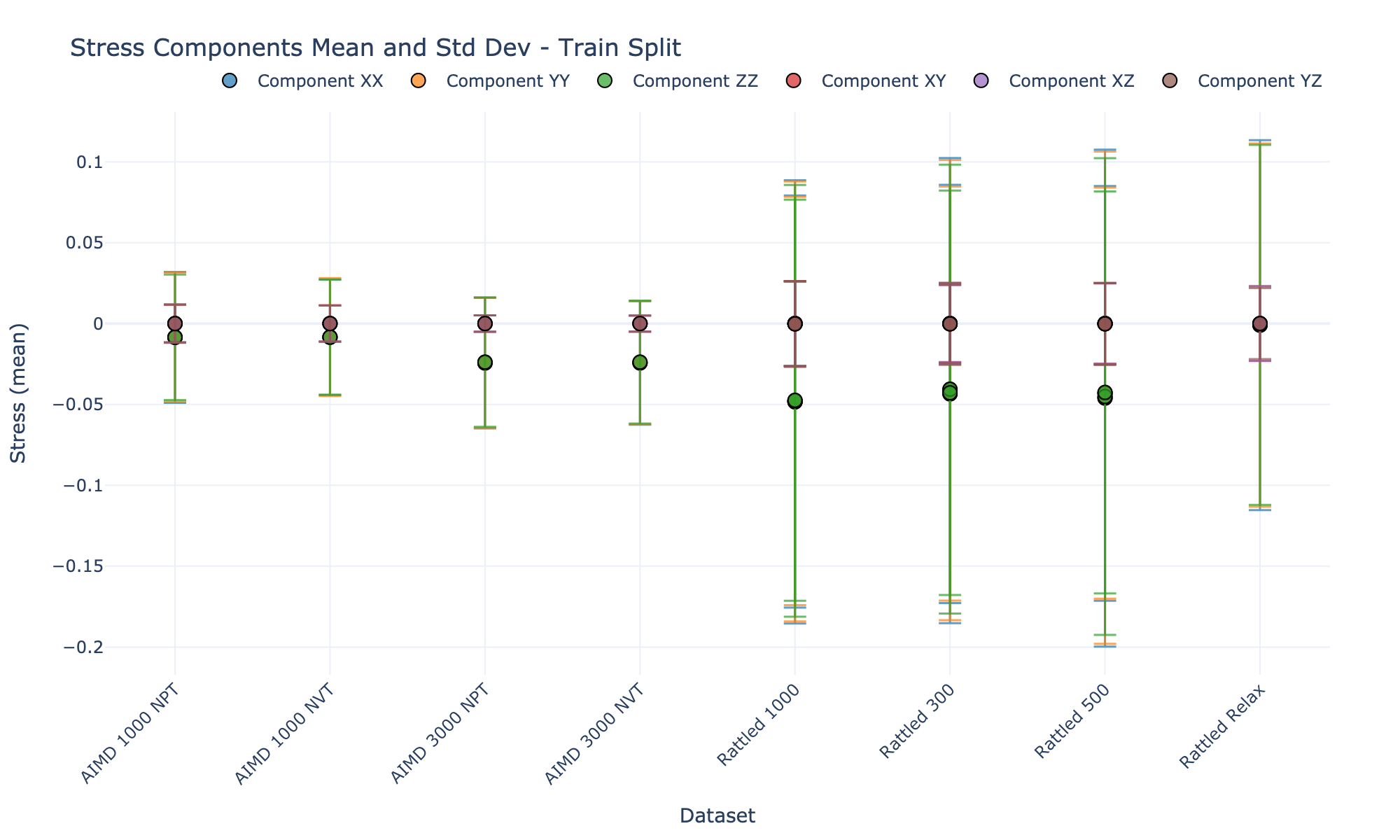}\includegraphics[scale=0.12]{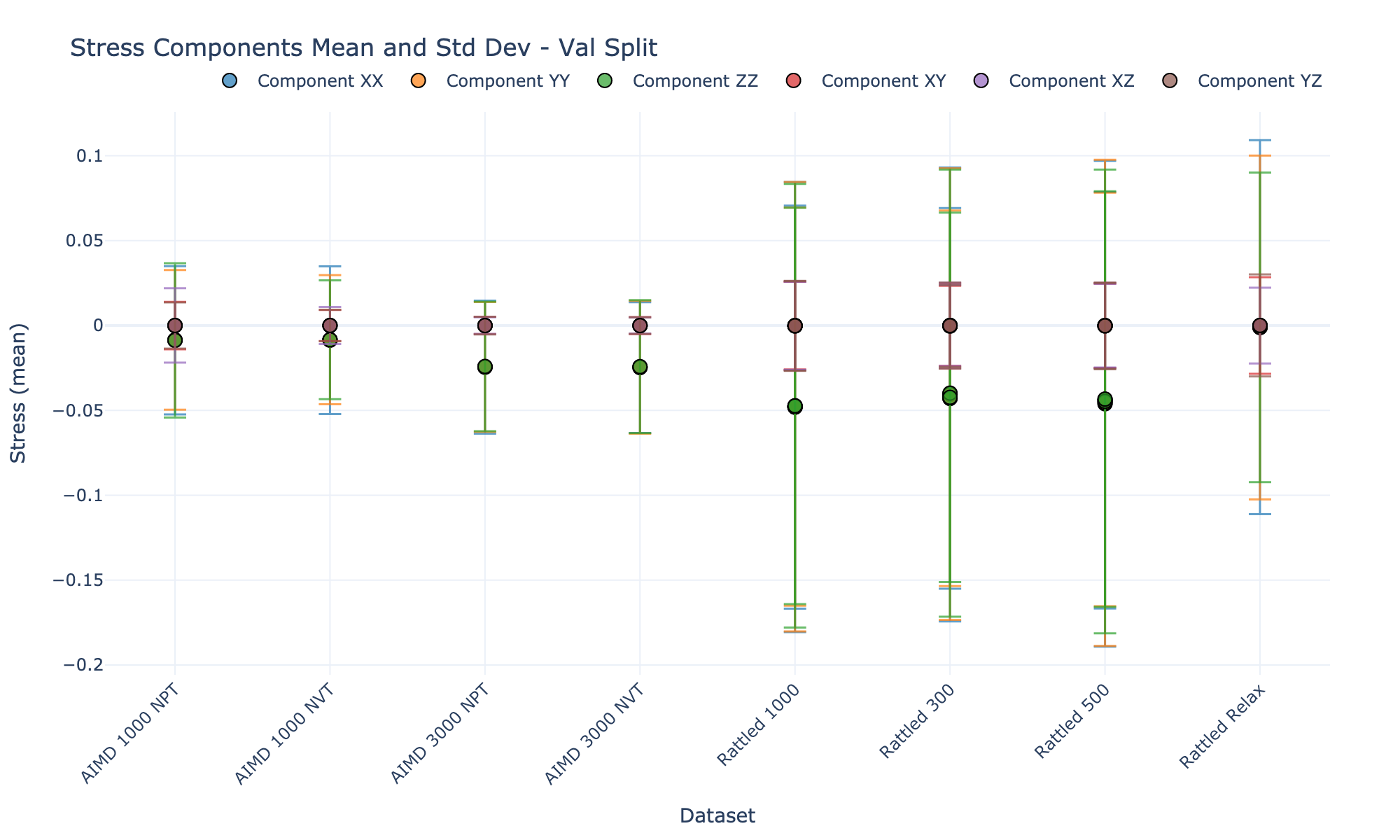}
\caption{We computed mean and standard deviations of stress components for datasets and splits.}
\label{fig:data_analysis_stress}
\end{figure}

\begin{figure}
\centering
\includegraphics[scale=0.12]{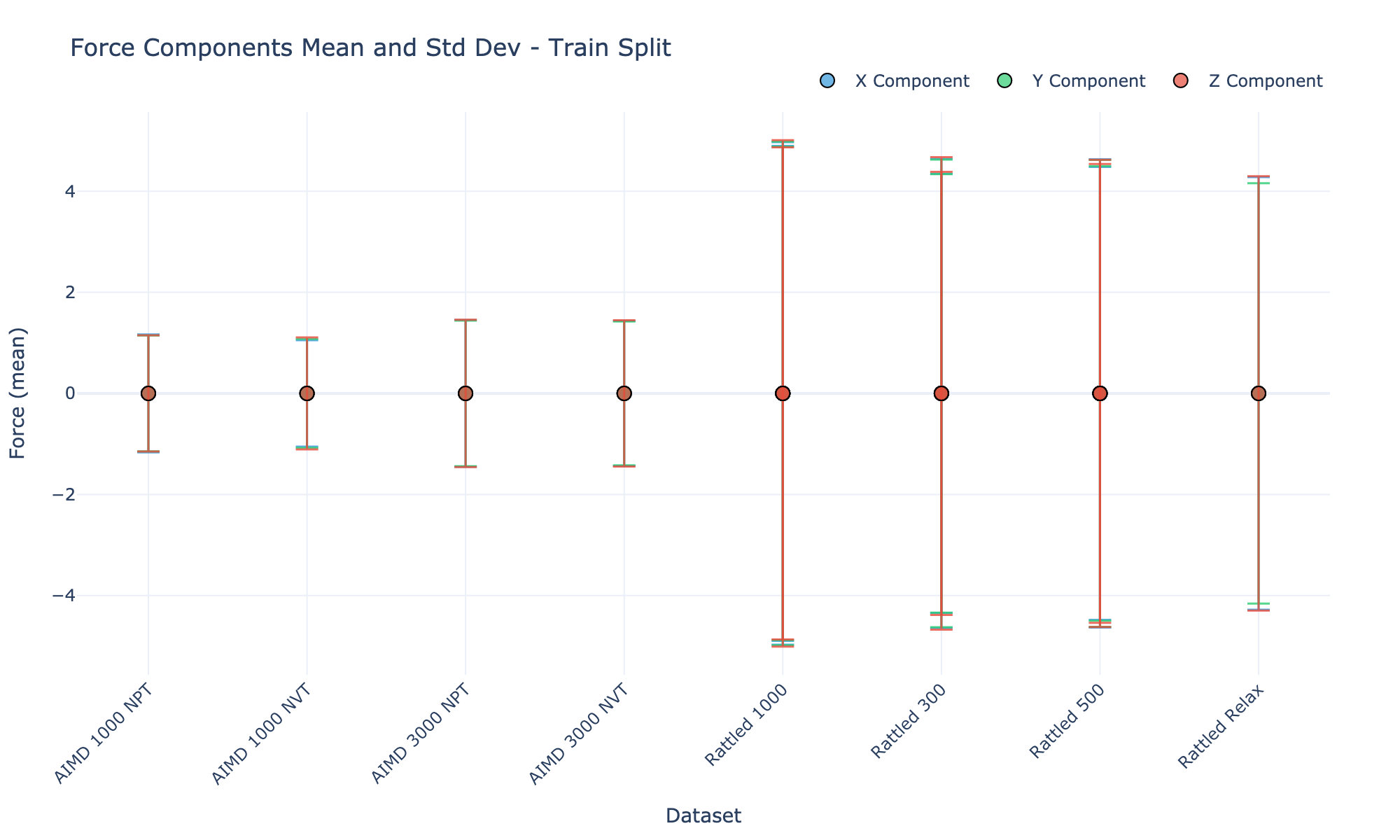}\includegraphics[scale=0.12]{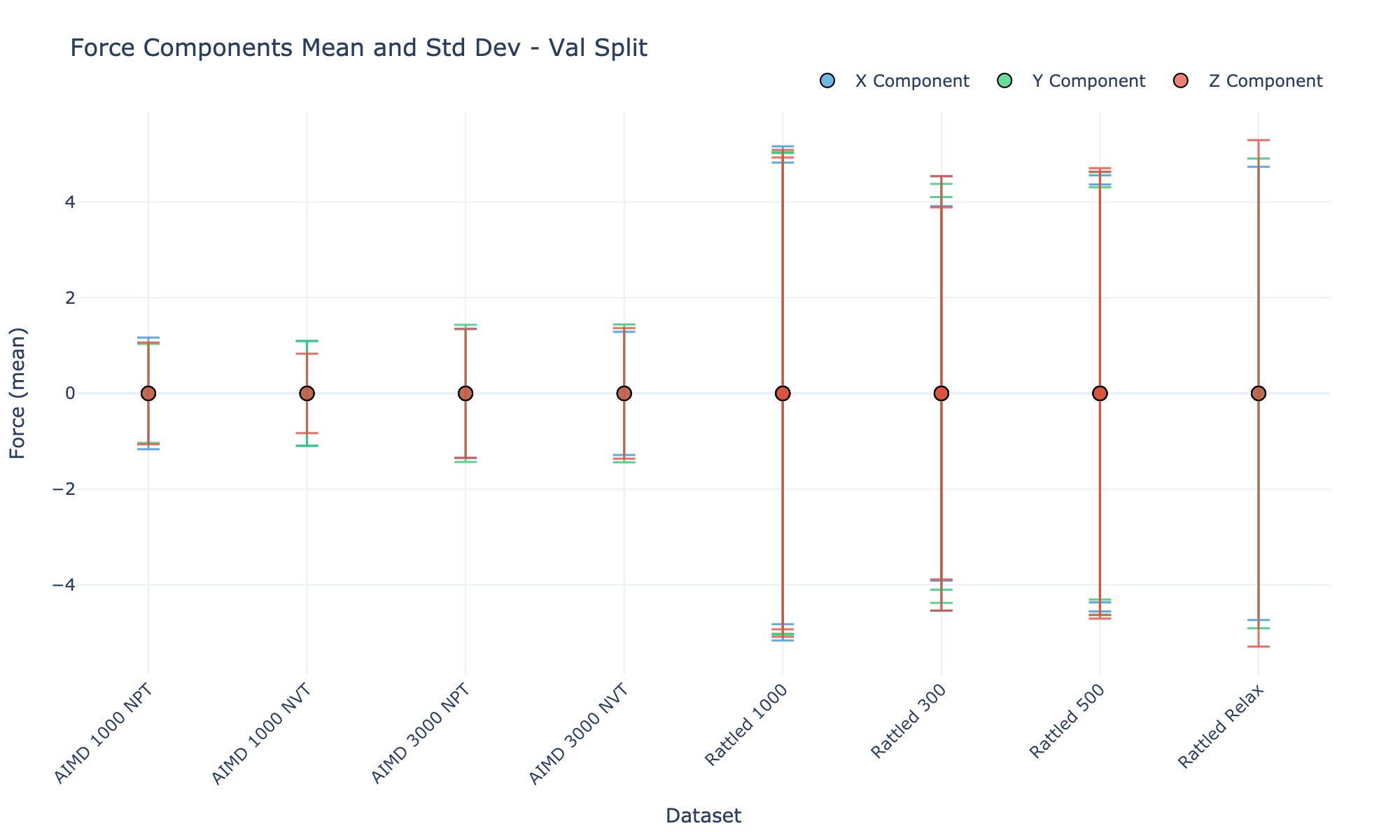}
\caption{We computed mean and standard deviations of force components for datasets and splits.}
\label{fig:data_analysis_force}
\end{figure}

We computed summary statistics of OMat24 datasets to better understand the data. As demonstrated in Figures \ref{fig:data_analysis_energy}, \ref{fig:data_analysis_stress}, and \ref{fig:data_analysis_force} the distributions of energy, stresses, and forces vary significantly from dataset to dataset within a split but are relatively consistent for the same dataset across splits.

\subsubsection{Compute}
To run our scaling experiments, we use GPU compute from a Savio cluster through the University of California, Berkeley. For each model, we separately vary the amount of training data, the amount of model parameters, and the amount of floating point operations (FLOPS) to empirically find scaling laws for performance. Overall, we create similar results to OpenAI's paper on the scaling of language modeling \cite{ScalingLawsForLLMs}, applied to the domain of materials data.

\subsubsection{Data Flow}
\begin{figure}[h]
\centering
\includegraphics[scale=1]{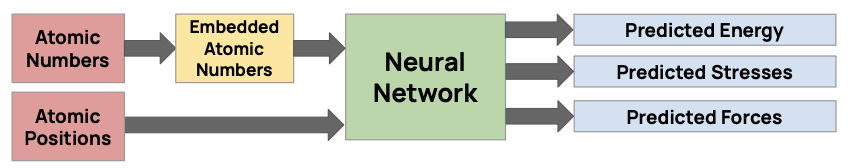}
\caption{Data flows in a multi-stage pipeline.}
\label{fig:data_flow}
\end{figure}

Figure \ref{fig:data_flow} demonstrates the flow of data through some of the model architectures. This multi-stage data pipeline allows models to fit predictions to OMat24 data. The inputted atomic data (red boxes) is embedded and concatenated (yellow box). We then input these results into the main neural network model being used, which is either a transformer or EquiformerV2, creating a rich representation of the input (green box). Finally, we apply two-layer FCNs including a SiLU nonlinearity to represent predicted outputs: energy, forces, and stresses (blue boxes).

The transformer architecture's embedding process implements several sophisticated encoding techniques to represent atomic systems effectively. In detail, the model employs three parallel embedding pathways that capture different aspects of the atomic structure:

First, atomic numbers are processed through a learned embedding layer (implemented as \texttt{atom\_type\_embedder = nn.Embedding(max\_num\_elements, d\_model)}), which maps each element to a dense vector of dimension \texttt{d\_model}. This approach allows the model to learn chemical similarities between elements rather than treating them as independent categorical variables.

Simultaneously, the model encodes spatial information through two complementary representations:
\begin{enumerate}
    \item Cartesian coordinates (absolute positions) are embedded using a two-layer MLP with a SiLU activation function. This captures the absolute atomic positions in 3D space.
    \item Fractional coordinates are similarly embedded through a separate two-layer MLP, providing information about relative positions within the unit cell---particularly valuable for crystalline materials.
\end{enumerate}

Additionally, the model incorporates position-aware representations using sinusoidal encodings based on atom indices. A function generates these embeddings using sine and cosine functions at different frequencies, following the approach introduced in the original Transformer paper but adapted for atomic systems. This positional information helps the model distinguish atom ordering within structures.

All these embeddings are summed (not concatenated as stated earlier) to form a comprehensive atom-level representation that preserves both chemical and spatial information. The resulting embeddings are then processed by the transformer encoder, where self-attention mechanisms allow each atom to attend to all other atoms in the structure, capturing complex many-body interactions regardless of spatial proximity.

\subsubsection{Loss}
A combination of the outputs needs to be fed into the models to measure performance via the loss function, which is a measure of how incorrect predictions are. The loss values backpropagate through our models, changing the models' weights, allowing the models to more accurately predict from data. Our loss function is the same as the formulation used in OMat24: a linear combination of individual losses for the energy, isotropic stresses, anisotropic stresses, and atomic forces \cite{OMat24}. This way, the models can predict all relevant types of material properties.

\subsubsection{Scaling}
Models with more training data, more parameters, and more compute are typically more capable of accurately predicting output data, so our team independently scales all three properties to generate scaling laws of how such scaling affects performance.

To scale model parameters, we choose different hyperparameters for the architectures of the models. For instance, we often increase model depth, allowing the models to further identify and express patterns in the inputs. We provide more detail in Appendix $4.1$.

\subsubsection{Training Settings}
Many of our training settings can be adjusted via the command line. These settings include but are not limited to batch size, number of epochs, maximum learning rate, fraction of training data, fraction of validation data, gradient clip, period of validation, period of visualization, distributed training, mixed precision, and data caching. Many of our numerical settings described below apply to EquiformerV2, not ADiT.

Batch size refers to the amount of data points that pass forward through a model at one given time. A larger batch size allows for increased throughput in training, which decreases training time. However, a batch size that is too large can exceed the memory capacity of the model, leading to a delicate balance between what is a helpful speedup and a limitation. We set our batch size to $32$.

The number of epochs is the amount of times that the model being trained sees each material. When a model is trained for more epochs, the model tends to predict that molecule more accurately. We set the number of epochs to $50$, but it is possible that a model could be trained for less epochs than this due to early stopping.

The learning rate represents the rate at which the model makes changes to its weights during backpropagation. We schedule the learning rate for our model in line with previous work: providing a learning rate `warmup' followed by a cosine decay. In particular, the initial learning rate is $20\%$ of the maximum learning rate of $6 \cdot 10^{-4}$, and the learning rate is increased until $1\%$ of the total maximum training steps. A cosine decay follows, with a minimum learning rate that is $1\%$ of the maximum learning rate \cite{OMat24}.

The training data and validation data fractions are the proportions of the datasets being used which will be used for training and validation. For example, if a dataset has $1$M data points and the training data proportion is set to $0.1$, the model being trained can learn from $100$k data points. We scale the amount of training data points, which greatly affects performance (see Figure \ref{fig:dataset_scaling1}). The amount of validation data points needs to be relatively large to make the validation data used be somewhat representative of the validation dataset. 

The gradient clip limits the amount that the model can update weights during training. We set the gradient clip to $100$.

The periods of validation and visualization determine how frequently the validation loss is captured and how frequently visualizations are rendered from. We set the validation period to $2$ epochs and the visualization period to $5$ epochs.

With distributed training, model training is done across multiple GPUs to increase throughput. In one experiment, we trained a transformer with roughly $11$M parameters on roughly $32$K materials with a batch size of $256$ via RTX$2000$ Ada GPUs. When trained on just one GPU, the model took $90.41$ seconds per epoch; when trained on seven GPUs, the model took $15.27$ seconds per epoch (an $83\%$ decrease). 

Mixed precision refers to the concept of storing different pieces of data with different amounts of floating-point precision. By reducing the floating-point precision, less information needs to be stored. This can decrease memory and time spent in training: when we trained a transformer with roughly $1$M parameters with a batch size of $256$, mixed precision reduced memory from $3,013$ MiB to $2,186$ MiB (a $27\%$ decrease) and reduced training time from $7.95$ seconds per epoch to $5.80$ seconds per epoch (a $27\%$ decrease).

Data caching, when enabled, allows training data to be loaded in once instead of having to do so every epoch. This can speed up training.

\subsubsection{Visualization}
\begin{figure}[h]
\centering
\includegraphics[scale=0.2]{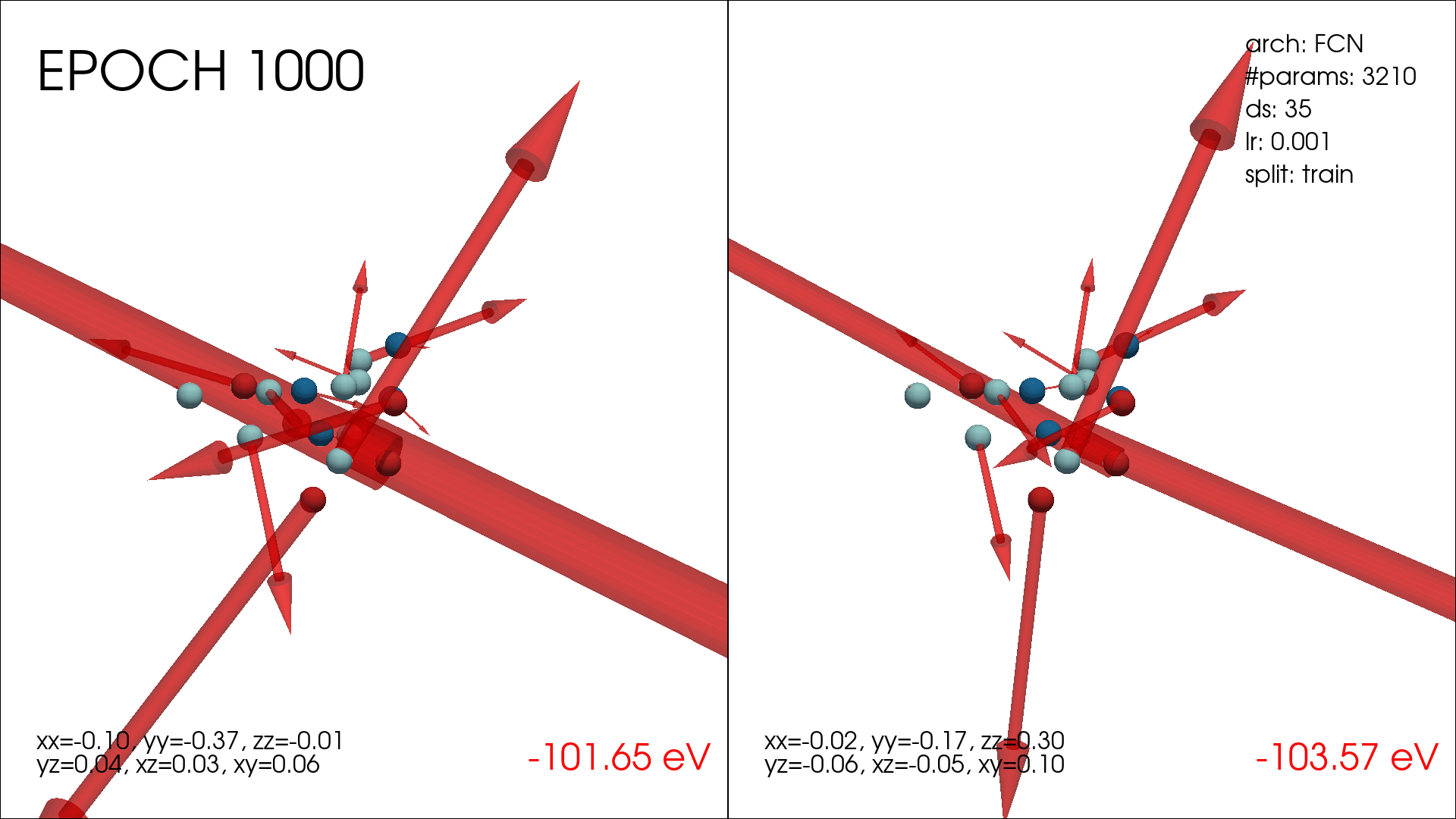}
\caption{We developed a system that renders visualizations of training. The left panel of the visualization represents actual properties of a material, while the right panel represents predicted properties of that material. The bottom-left of each panel denotes stresses; the bottom-right of each panel denotes energy; the red arrows in each panel denote atomic forces.}
\label{fig:visualization}
\end{figure}

To allow for interpretability of results, we also designed a program that visualizes the actual and predicted material properties: energy, stresses, and forces (see Figure \ref{fig:visualization}).

\subsubsection{Plots}
To present our results (subsection 3.2), we designed multiple types of plots. To scale training data (Figure \ref{fig:dataset_scaling1} and Figure \ref{fig:dataset_scaling2}), we trained a select few configurations of transformer and EquiformerV2 models for up to $50$ epochs on various amounts of training data. To scale model sizes (Figure \ref{fig:model_scaling1} and Figure \ref{fig:model_scaling2}), we trained configurations of the models for up to $50$ epochs. To scale compute (Figure \ref{fig:compute_scaling1} and Figure \ref{fig:compute_scaling2}), we tracked FLOPS when training the models to create compute curves for different model configurations.

\subsubsection{Inference}
We saved checkpoints of models from various epochs of training so that we can run inference on the models. This involves one forward pass through the model on an inputted material and computing resulting error metrics.

\subsubsection{Baseline Models}
To compare our performance for the deep learning models, we developed baseline models that we expect to perform less effectively than our best deep learning models. These naive baselines serve as benchmarks for performance.

\subsection{Empirical Results}
\subsubsection{Dataset Scaling Experiments}

\begin{figure}[h!]
\centering
\includegraphics[scale=0.325]{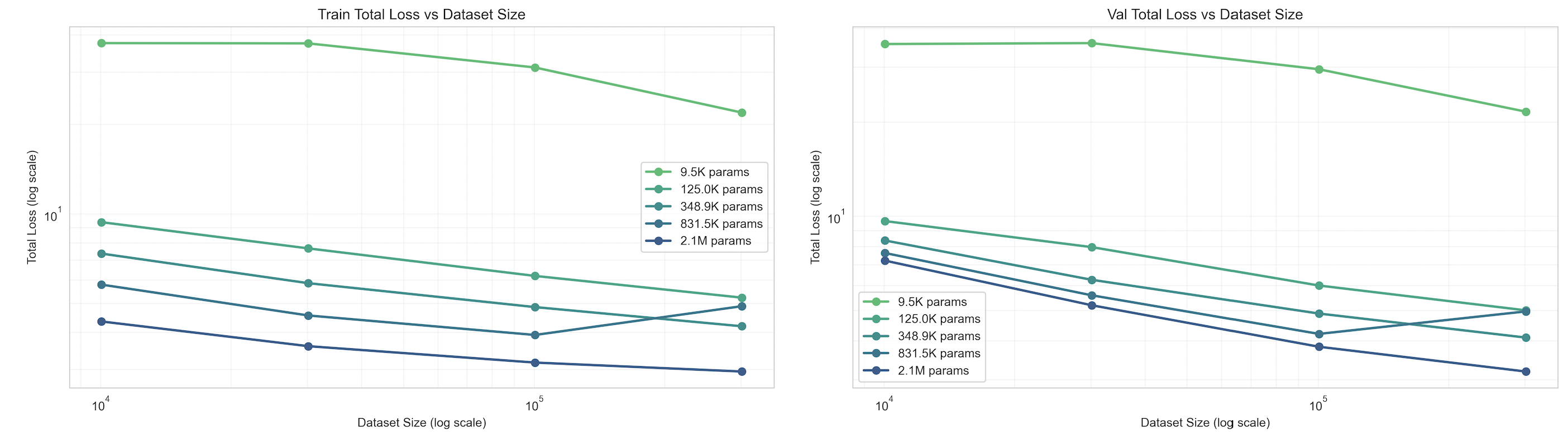}
\caption{Different amounts of training data yield different model size curves for the EquiformerV2.}
\label{fig:dataset_scaling1}
\end{figure}

Our experiments reveal distinct power-law scaling relationships between model performance and dataset size for the EquiformerV2. As shown in Figure~\ref{fig:dataset_scaling1}, both the EquiformerV2 models exhibit characteristic performance improvements with increased training data, following the form $L = \alpha \cdot D^{-\beta}$, with $D$ being the number of data points seen. The $2.1$M parameter architecture, which is our most representative one, scales according to $L \approx 64.7 \cdot D^{-0.242}$. Qualitatively, we also observe that as the dataset size increases, the validation loss values reduce regardless of model size, showing the architecture's capacity to successfully generalize from the data.

\begin{figure}[h!]
\centering
\includegraphics[scale=0.425]{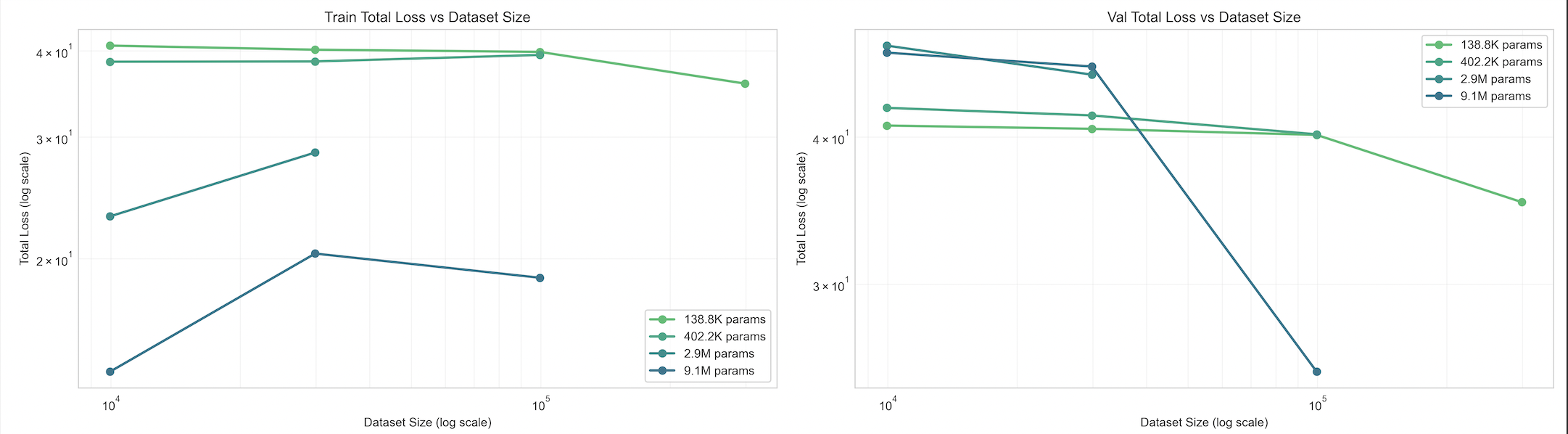}
\caption{Different amounts of training data yield different model size curves for the Transformer.}
\label{fig:dataset_scaling2}
\end{figure}

Figure~\ref{fig:dataset_scaling2} depicts a significantly different trend for the Transformer architecture. Since the transformers required $10$x more compute to reach sufficient loss values, we have insufficient results to demonstrate successful scaling. However, we see that in most of our models tested, the validation loss plateaus at $40$, without showing future signs of improvement. With increasing parameter counts to $2-9$M (compared to $2$M for the largest EquiformerV2 we trained), however, we are able to break past this barrier and achieve a lower loss. For the $2.9$M parameter transformer, we achieve a scaling law of $L \approx 77.1 \cdot N^{-0.052}$.

This pronounced difference in scaling behavior indicates that EquiformerV2 more effectively leverages additional training data, with performance improvements occurring at a faster rate than standard transformers. This underscores the architecture's superior ability to extract and leverage structural information from molecular data.

\subsubsection{Model Scaling Experiments}

\begin{figure}[h!]
\centering
\includegraphics[scale=0.325]{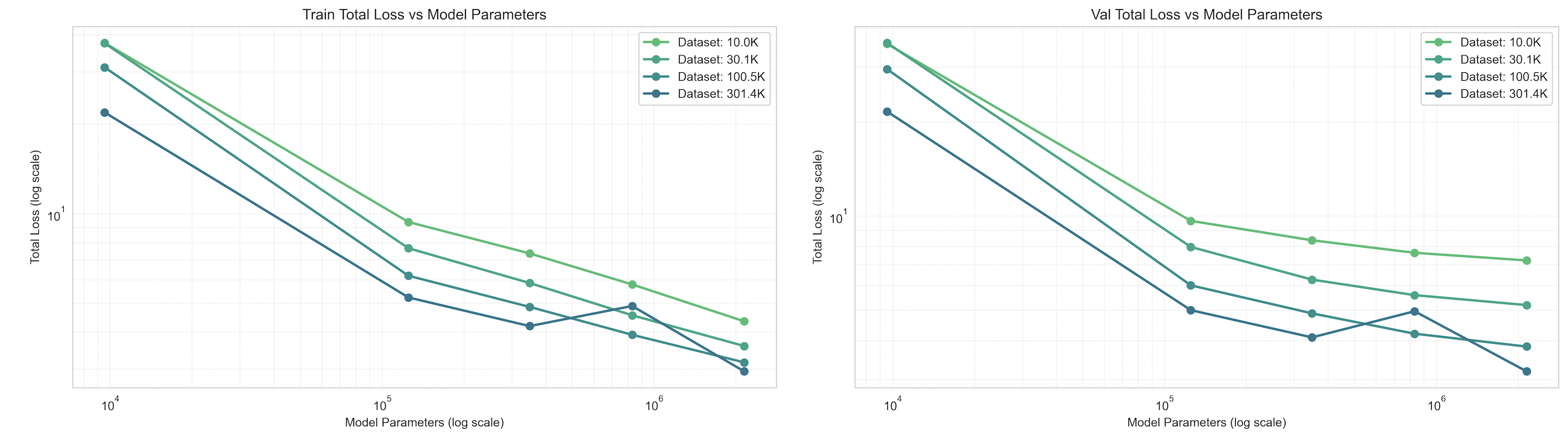}
\caption{Different model sizes yield different training data scaling curves for the EquiformerV2.}
\label{fig:model_scaling1}
\end{figure}

The next set of results we observe indicate a strong scaling behavior when increasing model sizes given a fixed amount of data. Similar to the previous section, our model scaling experiments reveal significant differences in how transformer and EquiformerV2 architectures respond to increased parameter counts. As shown in Figure~\ref{fig:model_scaling1}, 
the EquiformerV2 once again demonstrates a power-law scaling relationships with a linear trend on a log-log scale.

On $100$K data samples, we see the Equiformer models follow $L \approx 776 \cdot P^{-0.383}$, where $P$ is the number of non-embedding model parameters. There is a single anomaly for the $300$K dataset where both the train and validation losses for a particular model deviate from the trend. During our investigation, we can attribute this to a random bad case of model initialization.

\begin{figure}[h!]
\centering
\includegraphics[scale=0.425]{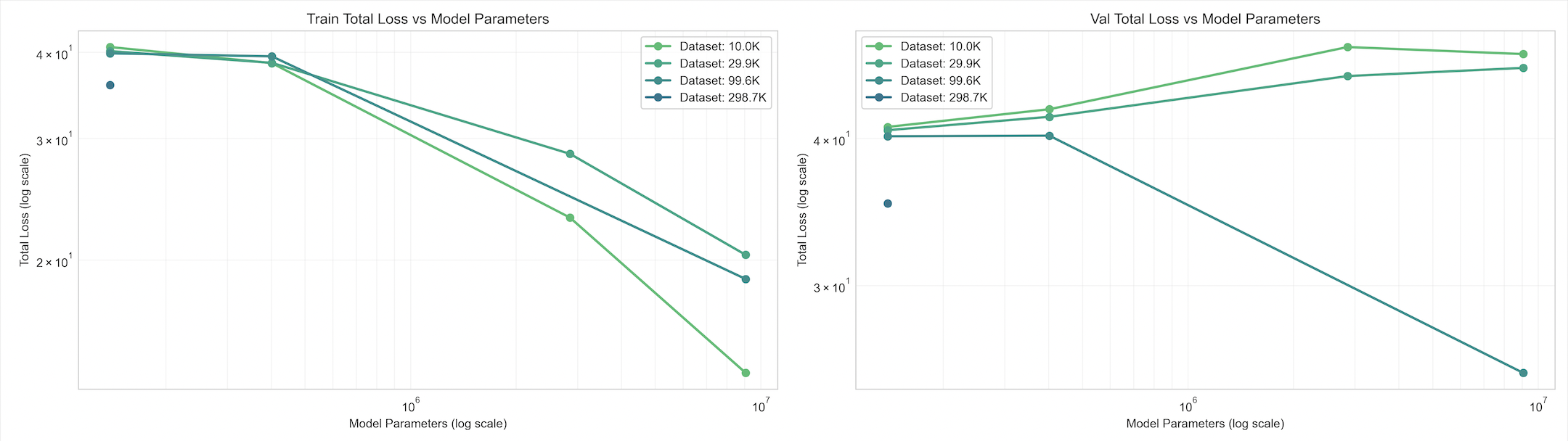}
\caption{Different model sizes yield different training data scaling curves for the Transformer.}
\label{fig:model_scaling2}
\end{figure}

Figure~\ref{fig:model_scaling2} illustrates how the Transformer architecture scales across different dataset sizes. The left panel shows that while the train loss follows a predictable trend, the models overfit on low amounts of data as seen by the validation loss on the right plot. We have insufficient data for the $300$K dataset size due to the compute restrictions mentioned in the section above, however the $100$K dataset gives us a scaling law of $L \approx 175 \cdot P^{-0.120}$.

EquiformerV2 models maintain substantially lower absolute loss values across all parameter ranges and dataset sizes, with even the smallest EquiformerV2 models approaching performance levels that require significantly larger transformer models. This highlights how architectural inductive biases allow more efficient parameter utilization for modeling molecular systems. The difference in scaling exponents indicates that EquiformerV2 more efficiently utilizes additional parameters, achieving greater performance improvements per parameter increase.

\subsubsection{Compute Scaling Experiments}

\begin{figure}[h]
\centering
\includegraphics[scale=0.45]{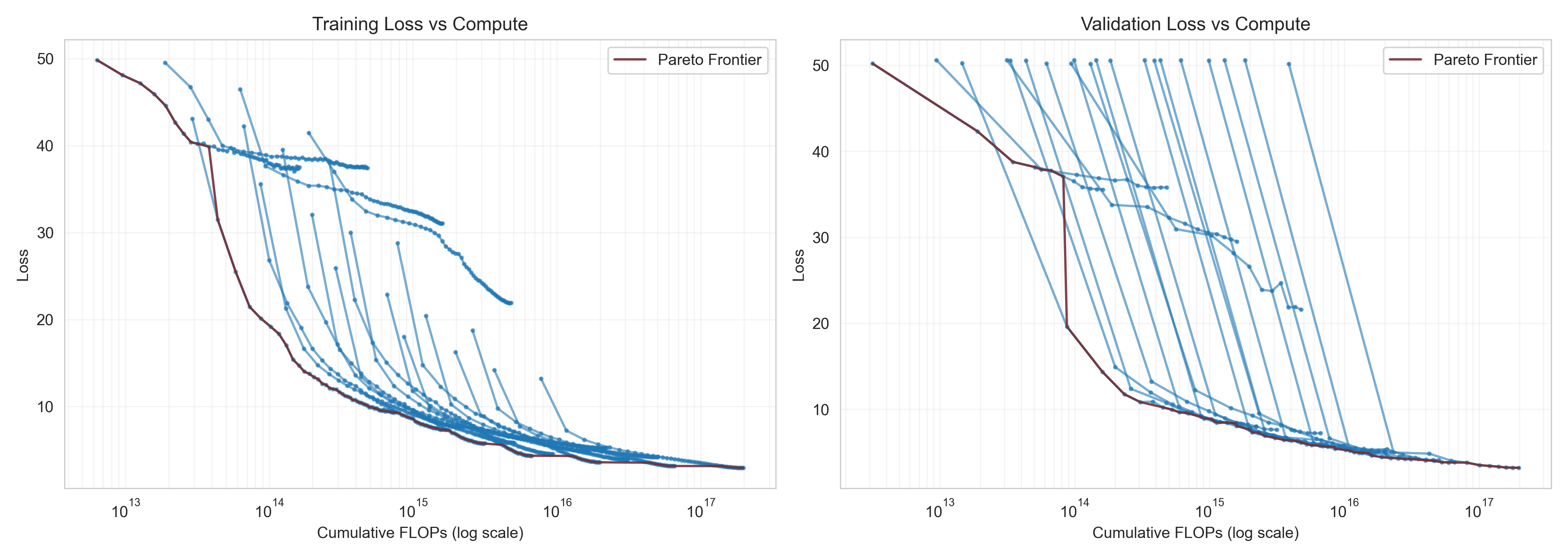}
\caption{We measure the effects of compute on loss for EquiformerV2.}
\label{fig:compute_scaling1}
\end{figure}

We conducted a comprehensive analysis of how model performance scales with computational resources, measuring compute in Floating Point Operations (FLOPs). Figure~\ref{fig:compute_scaling1}, shows the loss curves on both the train and validation data for the EquiformerV2 architecture. The Pareto frontier demonstrates a consistent power-law relationship between FLOPs and model performance, with loss values decreasing from approximately 50 to 3 across four orders of magnitude of compute ($10^{13}$ to $10^{17}$ FLOPs). Individual training trajectories show that larger models (requiring more FLOPs per epoch) achieve lower final losses but require substantially more compute to converge. The validation plot exhibits greater volatility than training, with numerous instances where validation loss temporarily increases during training before ultimately decreasing again. Notably, models operating below $10^{14}$ FLOPs struggle to achieve losses under 20, while the steepest improvements occur in the $10^{14}$ to $10^{15}$ FLOP range. At the highest compute regime of more than $10^{16}$ FLOPs, diminishing returns become evident: the Pareto frontier begins to flatten, with larger investments in computation yielding increasingly marginal improvements. Training and validation curves closely align at both the highest and lowest compute levels, with the most significant divergences occurring in the middle compute range.

\begin{figure}[h]
\centering
\includegraphics[scale=0.45]{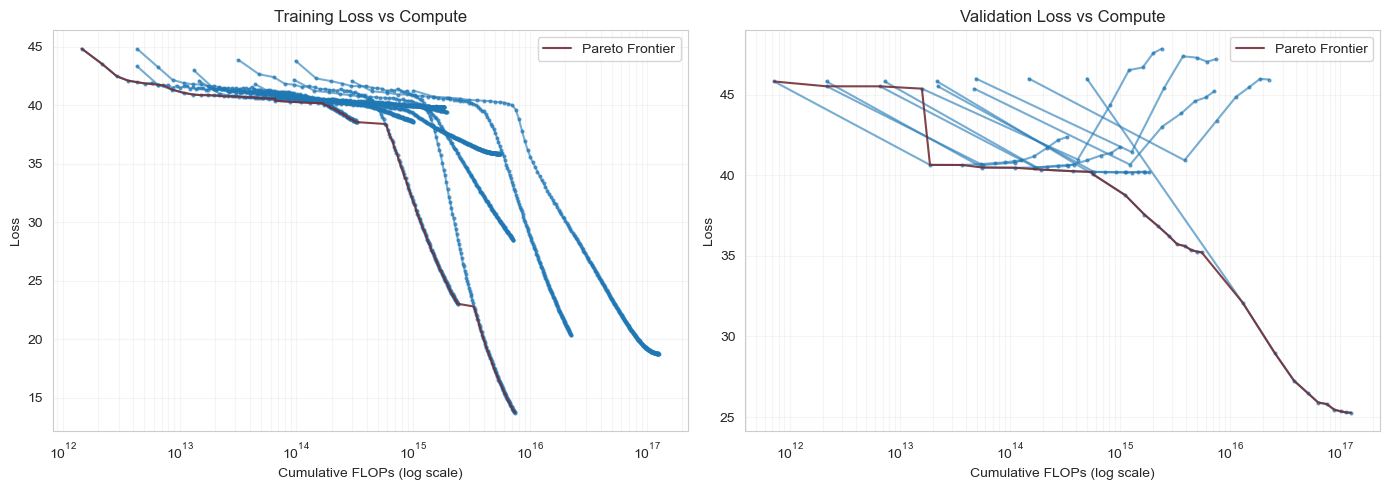}
\caption{We measure the effects of compute on loss for transformer.}
\label{fig:compute_scaling2}
\end{figure}

The compute scaling analysis for the Transformer in Figure~\ref{fig:compute_scaling2} reveals different efficiency patterns compared to EquiformerV2. Transformer models require much more FLOPs to begin their learning process. Most models, constrained by both parameters and the size of the dataset, plateau with a validation loss of $40$. Verifying the model output, these transformers achieve a loss of $40$ by setting the forces to $0$, regardless of their input, thereby demonstrating no actual learning. Only beyond the $10^{15}$ FLOP scale do we see that transformers demonstrate actual learning, with the lowest validation loss achieved around $25$. Hence, given the same amount of compute, we demonstrate that EquiformerV2 is a much more efficient architecture as it achieves a significantly lower validation loss of $3$.

Looking at the compute curves from a scaling perspective, the EquiformerV2 architecture exhibits clear power-law scaling relationships, with the validation loss decreasing smoothly after the first few epochs. The scaling law is fit by the equation $L \approx 4.99 \cdot 10^5 \cdot C^{-0.339}$. This fit improves if we remove the noise from the validation loss of the first few epochs, after which, the Pareto frontier can be seen to exhibit near linear decline on the log-log chart.

On the other hand, our experiments have insufficient data to derive fool-proof empirical compute scaling laws for Transformers on our material property prediction data. We do see potential of the models learning to generalize at the 100-300K data regime, which appears to be the main bottleneck in our experiments. Regardless of scaling dataset and model size to larger orders of magnitude, we are able to demonstrate that Transformers achieve a 8x higher (that is, worse) validation loss than the EquiformerV2, when comparing the two in a high compute regime.

\subsection{Discussion}
Our empirical analysis demonstrates clear, quantifiable scaling laws for neural network models in predicting material properties, highlighting significant differences between transformer and EquiformerV2 architectures. Notably, despite transformers achieving state-of-the-art results across many other domains—including natural language processing and computer vision—they consistently lag behind EquiformerV2 in material property prediction. This disparity emphasizes the unique advantage provided by explicitly incorporating physical symmetries, such as rotational and translational equivariance, into model architecture. Transformers, being less physically constrained, rely on implicitly learning these symmetries, which appears less effective at similar scales of data, model size, and compute.

These comparative findings serve as valuable benchmarks for researchers working at various computational scales. Future studies might push the boundaries further by evaluating model performance across additional magnitudes of scale, potentially into billions of parameters, to validate whether transformers can eventually approach EquiformerV2 performance through sheer size or if fundamental limitations persist. Moreover, exploring targeted data augmentation methods specifically tailored to enhance transformer performance could also provide meaningful improvements, potentially reducing the observed performance gap.

Additionally, examining other prominent graph neural network architectures, such as SchNet \cite{SchNet} or EScAIP \cite{EScAIP}, could provide further insight into whether the advantages of explicit physical symmetries are consistently reproducible across different equivariant modeling strategies. Such comparative studies would help clarify whether EquiformerV2's superior scaling behavior is uniquely effective or indicative of a broader, generalizable principle.

Lastly, expanding these scaling investigations into related material science tasks such as predicting reaction pathways, molecular dynamics trajectories, or properties of chemically more diverse datasets could greatly enrich our understanding of how generalizable these scaling laws are, thereby enhancing their practical utility in materials discovery.

\subsection{Conclusion}

Our experiments establish clear benchmarks for neural network scaling in materials science, demonstrating significant performance differences between transformers and EquiformerV2. The scaling laws reveal fundamental efficiency disparities across all dimensions examined. For dataset scaling, EquiformerV2 exhibits $L \approx 64.7 \cdot D^{-0.242}$, while transformers scale according to $L \approx 77.1 \cdot N^{-0.052}$. The nearly 5× larger scaling exponent ($-0.242$ versus $-0.052$) demonstrates EquiformerV2's dramatically superior ability to extract useful information from additional training data.

For parameter scaling on 100K samples, EquiformerV2 follows $L \approx 776 \cdot P^{-0.383}$, compared to transformers' $L \approx 175 \cdot P^{-0.120}$. The 3× larger scaling exponent ($-0.383$ vs $-0.120$) indicates that EquiformerV2 much more efficiently converts additional parameters into performance gains. The compute scaling analysis further reinforces this advantage, with EquiformerV2 demonstrating $L \approx 4.99 \cdot 10^5 \cdot C^{-0.339}$, while transformers show no meaningful learning below $10^{15}$ FLOPs and reach a validation loss floor approximately 8× higher than EquiformerV2.

These quantitative scaling laws highlight how architectural inductive biases that incorporate physical symmetries lead to substantially more efficient learning across all scaling dimensions. Future research could leverage these insights to further optimize materials discovery applications in domains like battery design, catalysis, and drug development.

\subsection{Code Availability}
Code for this project can be found at \url{https://github.com/akshaytrikha/materials-scaling}.



\bibliographystyle{IEEEtran}
\bibliography{sample}

\begin{thebibliography}{10}
\providecommand{\url}[1]{#1}
\csname url@samestyle\endcsname
\providecommand{\newblock}{\relax}
\providecommand{\bibinfo}[2]{#2}
\providecommand{\BIBentrySTDinterwordspacing}{\spaceskip=0pt\relax}
\providecommand{\BIBentryALTinterwordstretchfactor}{4}
\providecommand{\BIBentryALTinterwordspacing}{\spaceskip=\fontdimen2\font plus
\BIBentryALTinterwordstretchfactor\fontdimen3\font minus \fontdimen4\font\relax}
\providecommand{\BIBforeignlanguage}[2]{{%
\expandafter\ifx\csname l@#1\endcsname\relax
\typeout{** WARNING: IEEEtran.bst: No hyphenation pattern has been}%
\typeout{** loaded for the language `#1'. Using the pattern for}%
\typeout{** the default language instead.}%
\else
\language=\csname l@#1\endcsname
\fi
#2}}
\providecommand{\BIBdecl}{\relax}
\BIBdecl

\bibitem{OMat24}
\BIBentryALTinterwordspacing
L.~Barroso-Luque, M.~Shuaibi, X.~Fu, B.~M. Wood, M.~Dzamba, M.~Gao, A.~Rizvi, C.~L. Zitnick, and Z.~W. Ulissi, ``Open materials 2024 (omat24) inorganic materials dataset and models,'' 2024. [Online]. Available: \url{https://arxiv.org/abs/2410.12771}
\BIBentrySTDinterwordspacing

\bibitem{Alexandria}
J.~Schmidt, T.~F. Cerqueira, A.~H. Romero, A.~Loew, F.~J{\"a}ger, H.-C. Wang, S.~Botti, and M.~A. Marques, ``Improving machine-learning models in materials science through large datasets,'' \emph{Materials Today Physics}, p. 101560, 2024.

\bibitem{ADiT}
\BIBentryALTinterwordspacing
C.~K. Joshi, X.~Fu, Y.-L. Liao, V.~Gharakhanyan, B.~K. Miller, A.~Sriram, and Z.~W. Ulissi, ``All-atom diffusion transformers: Unified generative modelling of molecules and materials,'' 2025. [Online]. Available: \url{https://arxiv.org/abs/2503.03965}
\BIBentrySTDinterwordspacing

\bibitem{EquiformerV2}
Y.-L. Liao, B.~Wood, A.~Das, and T.~Smidt, ``Equiformerv2: Improved equivariant transformer for scaling to higher-degree representations,'' 2024.

\bibitem{BaiduScaling}
\BIBentryALTinterwordspacing
J.~Hestness, S.~Narang, N.~Ardalani, G.~F. Diamos, H.~Jun, H.~Kianinejad, M.~M.~A. Patwary, Y.~Yang, and Y.~Zhou, ``Deep learning scaling is predictable, empirically,'' \emph{CoRR}, vol. abs/1712.00409, 2017. [Online]. Available: \url{http://arxiv.org/abs/1712.00409}
\BIBentrySTDinterwordspacing

\bibitem{ScalingLawsForLLMs}
J.~Kaplan, S.~McCandlish, T.~Henighan, T.~B. Brown, B.~Chess, R.~Child, S.~Gray, A.~Radford, J.~Wu, and D.~Amodei, ``Scaling laws for neural language models,'' 2020.

\bibitem{NeuralMessagePassing}
J.~Gilmer, S.~S. Schoenholz, P.~F. Riley, O.~Vinyals, and G.~E. Dahl, ``Neural message passing for quantum chemistry,'' 2017.

\bibitem{SchNet}
K.~T. Schütt, P.-J. Kindermans, H.~E. Sauceda, S.~Chmiela, A.~Tkatchenko, and K.-R. Müller, ``Schnet: A continuous-filter convolutional neural network for modeling quantum interactions,'' 2017.

\bibitem{QM9}
\BIBentryALTinterwordspacing
R.~Ramakrishnan, P.~O. Dral, M.~Rupp, and O.~A. von Lilienfeld, ``Quantum chemistry structures and properties of 134 kilo molecules,'' \emph{Scientific Data}, vol.~1, p. 140022, 2014. [Online]. Available: \url{https://doi.org/10.1038/sdata.2014.22}
\BIBentrySTDinterwordspacing

\bibitem{NequIP}
S.~Batzner, A.~Musaelian, L.~Sun, and et~al., ``E(3)-equivariant graph neural networks for data-efficient and accurate interatomic potentials,'' \emph{Nature Communications}, vol.~13, p. 2453, 2022.

\bibitem{EScAIP}
\BIBentryALTinterwordspacing
E.~Qu and A.~S. Krishnapriyan, ``The importance of being scalable: Improving the speed and accuracy of neural network interatomic potentials across chemical domains,'' in \emph{The Thirty-eighth Annual Conference on Neural Information Processing Systems}, 2024. [Online]. Available: \url{https://openreview.net/forum?id=Y4mBaZu4vy}
\BIBentrySTDinterwordspacing

\end{thebibliography}

\end{document}